\pdfoutput=1

\documentclass[11pt]{article}

\usepackage[]{acl}

\usepackage{times}
\usepackage{latexsym}
\usepackage{float}
\usepackage{mathrsfs}
\usepackage{amsthm,amsmath,amssymb}
\usepackage{multirow}
\usepackage[ruled]{algorithm2e}
\usepackage{threeparttable}
\usepackage{array}
\usepackage{booktabs}
\usepackage{bm}
\usepackage{graphicx}
\usepackage{ amssymb }
\usepackage{wasysym}
\usepackage{xcolor}
\usepackage{soul}
\usepackage{ dsfont }

\usepackage[T1]{fontenc}

\usepackage[utf8]{inputenc}

\usepackage{microtype}

\title{ELECTRA is a Zero-Shot Learner, Too
}

\author{Shiwen Ni and Hung-Yu Kao$^*$ \\
	Department of Computer Science and Information Engineering
	\\
	National Cheng Kung University \\
	\texttt{P78083033@gs.ncku.edu.tw,hykao@mail.ncku.edu.tw}
}
\begin{document}
\maketitle
\begin{abstract}
Recently, for few-shot or even zero-shot learning, the new paradigm “\emph{pre-train, prompt, and predict}” has achieved remarkable achievements compared with the “\emph{pre-train, fine-tune}” paradigm. After the success of prompt-based GPT-3, a series of \textbf{masked language model} (MLM)-based (e.g., BERT, RoBERTa) prompt learning methods became popular and widely used. However, another efficient pre-trained discriminative model, ELECTRA, has probably been neglected. In this paper, we attempt to accomplish several NLP tasks in the zero-shot scenario using a novel our proposed \textbf{replaced token detection} (RTD)-based prompt learning method. Experimental results show that ELECTRA model based on RTD-prompt learning achieves surprisingly state-of-the-art zero-shot performance. Numerically, compared to MLM-RoBERTa$_{large}$ and MLM-BERT$_{large}$, our RTD-ELECTRA$_{large}$ has an average of about \textbf{8.4\%} and \textbf{13.7\%} improvement on all 15 tasks. Especially on the SST-2 task, our RTD-ELECTRA$_{large}$ achieves an astonishing \textbf{90.1\%} accuracy without any training data. Overall, compared to the pre-trained masked language models, the pre-trained replaced token detection model performs better in zero-shot learning. The source code is available at: \url{https://github.com/nishiwen1214/RTD-ELECTRA}\footnote{Timeline of our work: start in early November 2021, complete the first draft at the end of February, and submit the first draft to ACL ARR in March.}.
\end{abstract}

\section{Introduction}
Not long ago, the large-scale pre-trained GPT-3 \citep{floridi2020gpt,brown2020language} model made waves in the NLP community by showing amazing few-shot or even zero-shot learning performance in a series of language understanding tasks. People only need to give natural language prompts and a few task demonstrations, then GPT-3 can make accurate predictions without updating any weights of its underlying language model. However, it is worth noting that GPT-3 has 175B parameters. It needs a huge and powerful language model to work properly, making it unusable in many real-world scenarios.
\begin{figure}[t]
	\centering
	\includegraphics[width=0.96\linewidth]{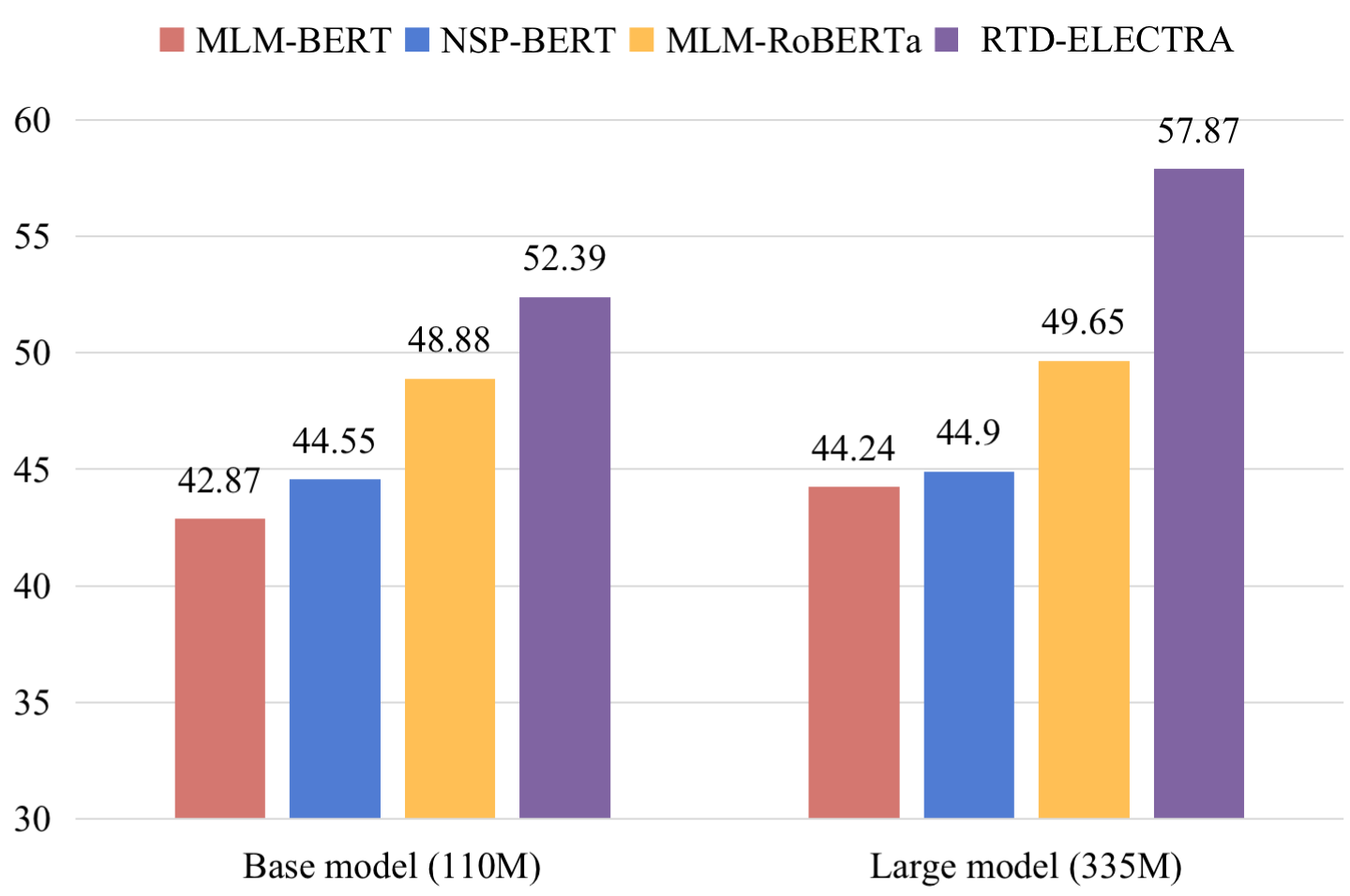}
	\caption{Average scores of various prompt-based zero-shot learning methods. 
	}
	\label{f1}
\end{figure}

Since then, some studies have made small language models such as BERT \citep{schick2021s,sun2021nsp}, ALBERT \citep{schick2021s}, and RoBERTa \citep{gao2021making} as few-shot and zero-shot learners by prompt-based learning, and achieved performance comparable to GPT-3 on a series of tasks. However, another efficient and powerful pre-trained language model, ELECTRA \citep{clark2020electra}, has been forgotten by people. As an alternative to masked language modeling (MLM), replaced token detection (RTD) was proposed in ELECTRA, a pre-training task in which the model learns to distinguish real input tokens from plausible but synthetically generated replacements. A vital advantage of the discriminative task is that the model learns from all input tokens instead of just the 15\% masked-out subset, making it more computationally efficient. Therefore, based on the excellent performance of RTD pre-training, we believe that the potential of ELECTRA can be further explored. In this paper, we associate RTD pre-training task with various downstream tasks by prompt-based learning, making ELECTRA a zero-shot learner.
As shown in Figure \ref{f1}, our prompt-based RTD-ELECTRA is much stronger than other prompt-based methods on base-sized and large-sized language models, which shows the tremendous potential of the ELECTRA model for zero-shot learning. 
The main contributions of this paper are summarized as follows:

\begin{itemize}
	\vspace{-1mm}
	\item To the best of our knowledge, we are the first to use ELECTRA as a zero-shot learner. Specifically, we propose an RTD-based prompt learning method applied to ELECTRA, making ELECTRA an excellent zero-shot and few-shot learner. 
	\vspace{-1mm}
	\item Experimental results show that our RTD-ELECTRA model achieves surprisingly state-of-the-art zero-shot performance on several public NLP datasets, which reveals that the potential of ELECTRA for zero-shot learning has been overlooked. 
	\vspace{-1mm}
	\item We further evaluate the few-shot learning ability of prompt-based RTD-ELECTRA, and the experimental results show that the few-shot learning ability of ELECTRA is also significantly higher than other popular masked language models (e.g., BERT, RoBERTa).
\end{itemize}
\section{Related work}
\subsection{Pre-trained Language Models}
\label{2}
With the advent of pre-trained language models, the field of NLP has ushered in rapid development and progress. \citet{devlin2019bert} first proposes the masked language model, BERT, and created the pre-train \& fine-tuning paradigm. The GPT proposed by \citep{radford2019language} is a left-to-right one-way language model based on Transformers, which uses a massive amount of corpus for pre-training. \citet{liu2019roberta} proposes modifications to the BERT pre-training procedure (e.g., the NSP pre-training task is removed) to improve end-task performance, called RoBERTa. ALBERT \citep{lan2019albert} uses a parameter sharing method to reduce the memory consumption of the original BERT model and replaces the next sentence prediction (NSP) task of the original BERT model with a new sentence-order prediction (SOP) task to improve the performance further. Subsequently, \citet{yang2019xlnet} proposes a Transformer-XL-based autoregressive pre-training model, XLNet, which learns bidirectional context via maximizing the expected likelihood of all permutations of the factorization order and overcomes the limitations of BERT due to its autoregressive formulation. \citet{clark2020electra} proposes a novel replaced token detection (RTD) pre-training task to improve the pre-training efficiency. After pre-training, the discriminator is used for downstream tasks. 

\begin{figure}[]
	\centering
	\includegraphics[width=0.99\linewidth]{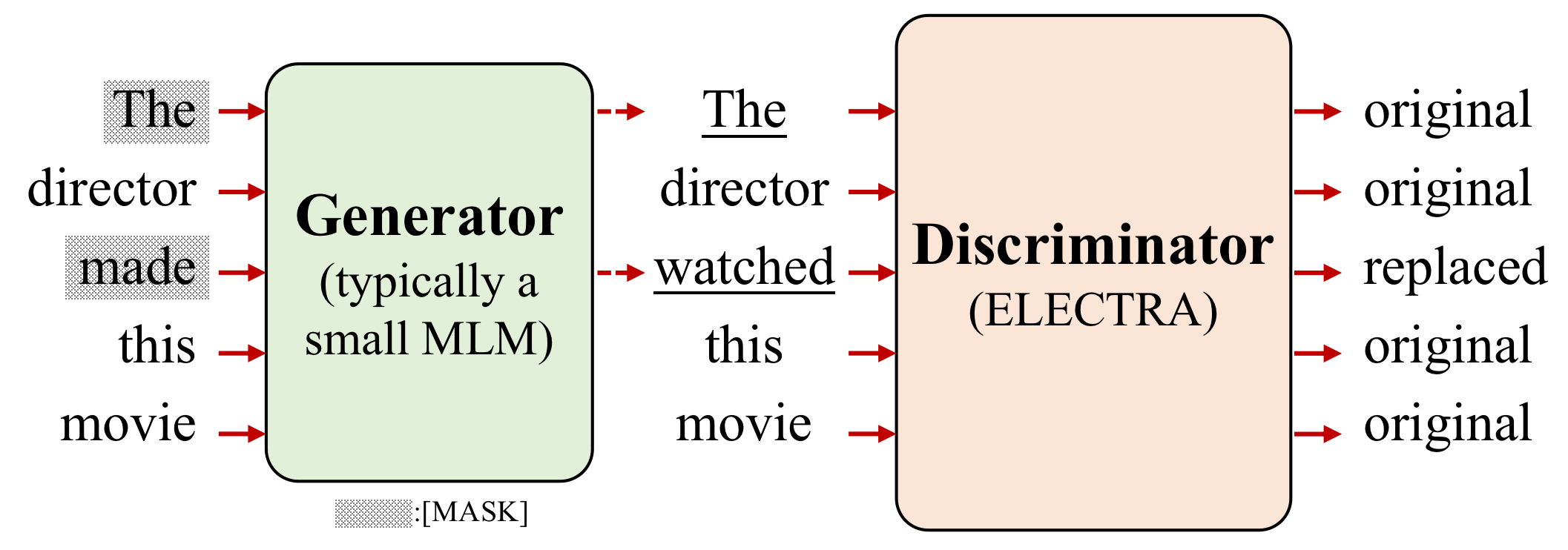}
	\caption{ An overview of \textbf{replaced token detection} (RTD) pre-training task \citep{clark2020electra}. 
	}
	\label{f2}
\end{figure}
\subsection{Prompt-based Learning}
Initially, a series of GPT-based studies \citep{radford2018improving,radford2019language,brown2020language} ushered in the era of language model prompting. The success of GPT-3 has led to the discovery that language models can be excellent zero-shot and few-shot learners with simple prompts \citep{liu2021pre}. \citet{schick2021s} performs prompt learning based on the MLM pre-training task and achieves surprising accuracy on a few samples. \citet{sun2021nsp} used a sentence-level pre-training task (NSP) for prompt learning to achieve promising results on zero-shot. \citet{gao2021making} proposes a simple and effective method for fine-tuning language models on few-shot, which automatically generates prompts and searches for the best prompts. \citet{liu2021gpt} employs trainable continuous prompt embeddings and improves both GPT and BERT well. Since then, many pieces of research \citep{jiang2020can,schick2021exploiting,he2021empirical} on prompt learning are based on MLM pre-training tasks, and there is no work on prompt learning based on RTD pre-training tasks. This work mainly focuses on what ELECTRA can achieve based on prompt learning without training data.

\begin{figure*}[t]
	\centering
	\includegraphics[width=0.99\linewidth]{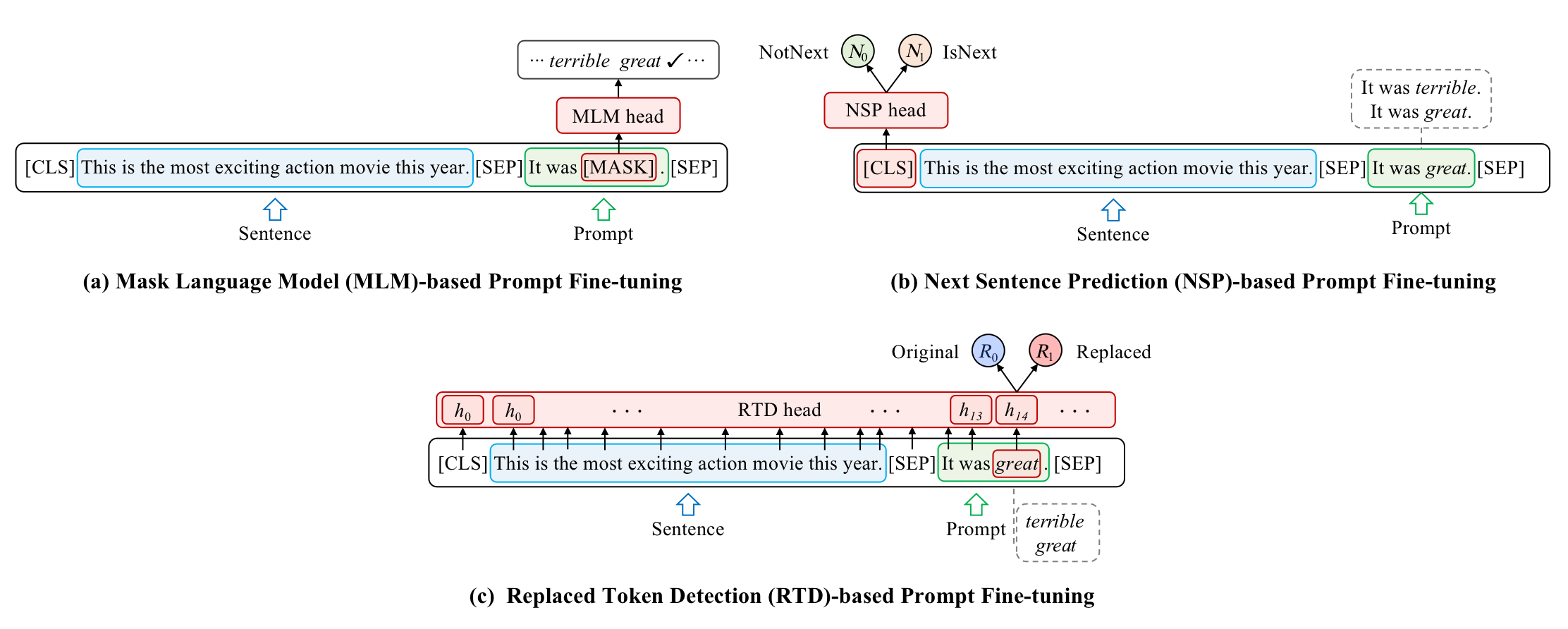}
	\caption{An illustration of (a) masked language model (MLM)-based prompt fine-tuning, (b) next sentence prediction (NSP)-based prompt fine-tuning, and (c) replaced token detection (RTD)-based prompt fine-tuning (our).
	}
	\label{f3}
\end{figure*}
\section{ELECTRA for Zero-Shot Learning}
This section will introduce how ELECTRA performs zero-shot learning based on prompts\footnote{Three concurrent works explore similar ideas to prompt ELECTRA for few-shot learning \citep{xia2022prompting,yao2022prompt,li2022pre}. \citet{yao2022prompt} and \citet{li2022pre} concatenate all the label words and input together for a forward pass, while we and  \citet{xia2022prompting} forward the input with different label words, respectively.}.
\subsection{Replaced Token Detection Pre-training}

We start by explaining the replaced token detection pre-training task. The overview of replaced token detection is shown in Figure \ref{f2}. This method trains two neural networks, a \textbf{generator} $G$ and a \textbf{discriminator} $D$. Both $G$ and $D$ are mainly composed of a transformer-based encoder. The encoder maps the sequence on the input token $\bm{x} = [x_1, ..., x_n]$ to the context vector representation sequence $\bm{h}(x) = [h_1, ..., h_n].$ For a given position $t$, (the example in Figure \ref{f2} only has the position of $x_t = ${\tt [mask]}, e.g., "The", "made"), the generator outputs the probability of using the softmax layer to generate a specific token $x_t$:
\begin{equation}\label{key}
p_G(x_t|\bm{x})=\frac{exp({\bm e}(x_t)^T{\bm h}_G({\bm x})_t)}{ \sum_{x^{'}} exp({\bm e}(x^{'})^T{\bm h}_G(\bm{x})_t)}
\end{equation} 
where $e$ denotes token embeddings. For a given position $t$, the discriminator predicts whether the token $x_t$ is “{\tt replaced}”, i.e., that it comes from the real data rather than the generator distribution, with a sigmoid output layer:
\begin{equation}\label{key}
D(\bm{x},t)={\rm sigmoid}({\bm w}^T{\bm h}_D(x)_t)
\end{equation}

More specifically, the generator is trained based on the MLM task. For a given input $\bm{x} = [x_1, x_2, ..., x_n]$, a set of random positions (integers between 1 and n) is first generated to mask $m = [m_1, ..., m_k]~(k = 15\% * n)$. The tokens in the selected location are replaced with [MASK] tokens, and the formula is as follows:
\begin{equation}\label{key}
\begin{split}
&x^{masked}={\rm REPLACE} (\bm{x},m,{\rm [MASK]}) \\
&s.t.~m_{i}\sim unif\{1,n\}~\rm{for}~i=1~\rm{to}~k.
\end{split}
\end{equation}
The generator learns to predict the original word of the masked token, and then generates a new word to replace the [MASK], the formula is as follows:
\begin{equation}\label{key}
\begin{split}
&x^{corrupt}={\rm REPLACE} (\bm{x},m,x^{G}), \\
&s.t.~x_{i}^{G}\sim P_G\{x_i|x^{masked}\}~{\rm for}~i~\in~m
\end{split}
\end{equation}

It’s worth noting that if the generator happens to generate the correct token, the token is considered "{\tt original}" and not "{\tt replaced}". For example, in Figure \ref{f2}, the words "The" to "[MASK]" and then to "The", the final discriminator output is "{\tt original}". The discriminator is trained to distinguish tokens in the data from tokens that have been replaced by generator samples.
During optimization, the loss functions of generator $G$ and a discriminator $D$ are:
\vspace{-1em}
\begin{equation}\label{key}
\mathcal{L}_{MLM}(\bm{x},\theta_G)= \mathds{E}(\sum_{i \in m}-logP_G(x_i|\bm{x}^{masked}))
\end{equation}
\vspace{-2.5em}
\begin{equation}\label{key}
\begin{split}
&\mathcal{L}_{Disc}(\bm{x},\theta_D)=\mathds{E}(\sum_{t =1}^{n}-\mathds{1}(x_{t}^{corrupt}=x_{t})\\&logD(x^{corrupt},t)-\mathds{1}(x_{t}^{corrupt}\neq \\& x_{t}log(1-D(\bm{x}^{corrupt},t)))
\end{split}
\end{equation}

The final optimization goal is:
\begin{equation}\label{key}
\mathop{{\rm min}}\limits_{\theta_G,\theta_D}\sum_{\bm{x}\in\mathcal{X}}\mathcal{L}_{MLM}(\bm{x},\theta_G)+\mathcal{L}_{Disc}(\bm{x},\theta_D)
\end{equation}

Generator $G$ is trained with maximum likelihood instead of adversarial training to fool the discriminator. After pre-training, we choose the discriminator (ELECTRA) for various downstream tasks.

\begin{table*}[]
	\centering
	\small 
	\renewcommand\arraystretch{1.2}
	\setlength\tabcolsep{5.pt}
	\begin{tabular}{lll}
		\toprule
		\textbf{Task} &  \textbf{Template (Prompt)} & \textbf{Label words} \\ \hline
		SST-2 & \textless{}CLS\textgreater{} {[}Input{]} \textless{}SEP\textgreater{} \underline{ This movie is {[}label{]}!! } \textless{}SEP\textgreater{} & \begin{tabular}[c]{@{}l@{}} great; terrible\end{tabular} \\ \hline
		SST-5 & \textless{}CLS\textgreater \underline{ This movie is {[}label{]}. } \textless{}SEP\textgreater{} {[}Input{]} \textless{}SEP\textgreater{} & \begin{tabular}[c]{@{}l@{}} perfect; good; okay;\\ bad; terrible\end{tabular} \\ \hline
		MR & \textless{}CLS\textgreater{}\underline{ It was {[}label{]}! } \textless{}SEP\textgreater{} {[}Input{]} \textless{}SEP\textgreater{} & \begin{tabular}[c]{@{}l@{}}great; terrible\end{tabular} \\ \hline
		CR & \textless{}CLS\textgreater{} {[}Input{]} \textless{}SEP\textgreater{} \underline{ I really {[}label{]} this product. } \textless{}SEP\textgreater{} & \begin{tabular}[c]{@{}l@{}} love; hate\end{tabular} \\ \hline
		MPQA & \textless{}CLS\textgreater{} {[}Input{]} \textless{}SEP\textgreater{} \underline{ {[}label{]} good, } \textless{}SEP\textgreater{} & \begin{tabular}[c]{@{}l@{}} really; not\end{tabular} \\ \hline
		Subj & \textless{}CLS\textgreater{} \underline{ {[}label{]} speaking. } \textless{}SEP\textgreater{} {[}Input{]} \textless{}SEP\textgreater{} & \begin{tabular}[c]{@{}l@{}}Subjectively; Objectively\end{tabular} \\ \hline
		TREC & \textless{}CLS\textgreater{} \underline{ The answer is about a {[}label{]}, } \textless{}SEP\textgreater{} {[}Question{]} \textless{}SEP\textgreater{} & \begin{tabular}[c]{@{}l@{}}definition; entity; meaning; \\person; place; number\end{tabular} \\ \hline
		CoLA & \textless{}CLS\textgreater{} \underline{ The grammar of the following sentence is {[}label{]}, }\textless{}SEP\textgreater{} {[}Input{]} \textless{}SEP\textgreater{} & \begin{tabular}[c]{@{}l@{}}correct; wrong\end{tabular} \\ \hline
		MNLI & \textless{}CLS\textgreater{} {[}premise{]} \textless{}SEP\textgreater\underline{ ? {[}label{]}, }{[}hypothesis{]} \textless{}SEP\textgreater{} & \begin{tabular}[c]{@{}l@{}} Yes; Maybe; No\end{tabular} \\ \hline
		SNLI & \textless{}CLS\textgreater{} {[}premise{]} \textless{}SEP\textgreater \underline{ ? {[}label{]}, }{[}hypothesis{]} \textless{}SEP\textgreater{} & \begin{tabular}[c]{@{}l@{}}Yes; Maybe; No\end{tabular} \\ \hline
		QNLI & \textless{}CLS\textgreater{} {[}premise{]} \textless{}SEP\textgreater\underline{ ? {[}label{]}! }{[}hypothesis{]} \textless{}SEP\textgreater{} & \begin{tabular}[c]{@{}l@{}}Yes; No\end{tabular} \\ \hline
		RTE & \textless{}CLS\textgreater{} {[}premise{]} \textless{}SEP\textgreater \underline{ ? {[}label{]}! } {[}hypothesis{]} \textless{}SEP\textgreater{} & \begin{tabular}[c]{@{}l@{}}Yes; No\end{tabular} \\ \hline
		MRPC & \textless{}CLS\textgreater{} {[}Question1{]} \textless{}SEP\textgreater \underline{ ? {[}label{]}, }{[}Question2{]} \textless{}SEP\textgreater{} & \begin{tabular}[c]{@{}l@{}}Yes; No\end{tabular} \\ \hline
		QQP & \textless{}CLS\textgreater{} {[}Question1{]} \textless{}SEP\textgreater \underline{ ? {[}label{]}, }{[}Question2{]} \textless{}SEP\textgreater{} & \begin{tabular}[c]{@{}l@{}}Yes; No\end{tabular} \\ \hline
		STS-B & \textless{}CLS\textgreater{} {[}Sentence1{]} \textless{}SEP\textgreater{}\underline{ ? NO!! } {[}Sentence2{]} \textless{}SEP\textgreater{} & Regression Task (0 $\sim$ 5) \\
		\bottomrule
	\end{tabular}
	\caption{Templates and label words that we used in experiments. The \underline{underlined} text is the task-specific prompt.}
	\label{t1}
\end{table*}
\begin{table*}[t]
	\renewcommand\arraystretch{1.2}
	\setlength\tabcolsep{7.5pt}
	\small
	\begin{tabular}{l|cccccccc|c}
		\toprule[1pt]
		\multirow{3}{*}{Methods} & \multicolumn{8}{c|}{\textbf{Single-sentence tasks}} & \multirow{3}{*}{AVG.} \\ \cline{2-9}&\textbf{SST-2} &\textbf{SST-5} &\textbf{MR} &\textbf{CR} &\textbf{MPQA} &\textbf{Subj} &\textbf{TREC} &\textbf{CoLA} \\ & (acc.)&(acc.)&(acc.)&(acc.)&(acc.)&(acc.)&(acc.)&(matt.)  \\ \hline 
		Majority class & 50.9 & 23.1 & 50.0 & 50.0 & 50.0 & 50.0 & 18.8 & 0.0& 36.60 \\ \hline
		\multicolumn{10}{c}{Prompt-based zero-shot learning (Base-sized models)} \\ \hline
		MLM-BERT&66.1&22.6&61.7&67.0&57.6&50.0&52.0&-5.6& 46.43\\
		NSP-BERT & 73.6 & 26.7 & 71.9 & 79.2 & \textbf{74.9} & 64.5 & 50.0 & -9.2 & 53.95 \\
		MLM-RoBERTa &77.8&\textbf{30.3}&77.7&\textbf{79.8}&70.1&53.6&27.4&\textbf{-3.61}&51.64 \\
		RTD-ELECTRA & \textbf{81.0} & 29.8 & \textbf{85.2} &79.0 & 61.7 & \textbf{61.2} & \textbf{57.0} & -5.14 & \textbf{56.22} \\ \hline 
		\multicolumn{10}{c}{Prompt-based zero-shot learning (Large-sized models)} \\ \hline
		MLM-BERT&60.1&22.8&54.3&64.4&57.9&50.4&51.6&2.3&45.48\\
		NSP-BERT & 74.9 & 24.7 & 76.2 & 76.8 & \textbf{75.4} & 49.3 & 48.6 & -5.6 & 52.54 \\
		MLM-RoBERTa & 83.6 & \textbf{35.0} & 80.8 & 79.5 & 67.6 & 51.4 & 32.0 & 2.0 & 53.99 \\
		RTD-ELECTRA & \textbf{90.1} & 34.9 & \textbf{84.5} &\textbf{79.7 }& 68.2 & \textbf{69.8} & \textbf{62.2} & \textbf{10.2} & \textbf{62.45} \\ \hline \hline
		
		\multirow{3}{*}{Methods} & \multicolumn{8}{c|}{\textbf{Sentence-pair tasks}} & \multirow{3}{*}{AVG.} \\ \cline{2-9}&\textbf{MNLI} &\textbf{MNLI} &\textbf{SNLI} &\textbf{QNLI} &\textbf{RTE} &\textbf{MRPC} &\textbf{QQP} &\textbf{STS-B} \\ & (m-acc.)&(mm-acc.)&(acc.)&(acc.)&(acc.)&(F1)&(F1)&(Pear.)  \\ \hline 
		
		Majority class & 32.7 & 33.0 & 33.8 & 49.5 & 52.7 & 81.2 & 0.0& - &35.36 \\ \hline
		\multicolumn{10}{c}{Prompt-based zero-shot learning (Base-sized models)} \\ \hline
		MLM-BERT&42.8&44.5&49.2&49.8&49.8&60.9&28.1&-10.6&39.31	 \\
		NSP-BERT & 31.2 & 31.9 & 35.4 & 45.9 & 46.9 & 26.3 & 44.9 & \textbf{18.7} & 35.15 \\
		MLM-RoBERTa &48.1&49.1&48.8&50.5&53.4&53.0&51.6&14.5& 46.12 \\
		RTD-ELECTRA & \textbf{52.6} & \textbf{51.8} & \textbf{55.5} & \textbf{50.1} & \textbf{55.2} & \textbf{75.7} & \textbf{54.3} & -6.7 & \textbf{48.56} \\ \hline 
		\multicolumn{10}{c}{Prompt-based zero-shot learning (Large-sized models)} \\ \hline
		MLM-BERT&46.6&47.9&49.2&49.9&53.1&59.5&37.2&0.6&43.00	 \\
		NSP-BERT & 37.4 & 38.2 & 42.7 & 42.5 & 52.7 & 24.4 & 23.3 & \textbf{17.0} & 37.26 \\
		MLM-RoBERTa & 50.8 & 51.7 & 49.5 & 50.8 & 51.3 & 61.9 & 49.7 & -3.2 & 45.31 \\
		RTD-ELECTRA & \textbf{56.1} & \textbf{55.9} & \textbf{56.8} & \textbf{57.1} & \textbf{64.3} & \textbf{70.0} & \textbf{51.5} & 14.6 & \textbf{53.29} \\ 
		\bottomrule[1pt]
	\end{tabular}
	\caption{Zero-shot experiment results. The results of NSP-BERT \citep{sun2021nsp} and MLM-RoBERTa \citep{gao2021making} are run with official code and default parameters (for sentence pair tasks, since the original NSP-BERT uses dev set, for fairness, we use the same prompt as RTD-ELECTRA to perform zero-shot experiments).}
	\label{t2}
\end{table*}
\subsection{RTD-based Prompt and Predict}
Our RTD-based Prompt learning need to build appropriate templates (prompts) for various tasks like other prompt-based learning methods. Since zero-shot learning does not rely on the training data of any downstream tasks, the constructed template must match the downstream task to the original pre-training task as much as possible. Different templates (prompts) and prediction strategies are required for various pre-training tasks. Figure \ref{f3} shows the prompt based on three different pre-training tasks. The downstream tasks can also be transformed into the RTD binary classification task ({\tt original/replaced}) through a specially designed template (prompt). For example, for the movie review binary classification task in Figure \ref{f3}, the input sentence is “\textit{This is the most exciting movie of the year},” the prompt is “\textit{It was [label word]},” and the label words are “\textit{great}” and “\textit{terrible}” respectively. If the model predicts the token "\textit{great}" to be \texttt{original}, then the input is \textbf{positive}, otherwise the model predicts "\textit{terrible}" to be \texttt{original}, then the input is \textbf{negative}. Next, we explain how to formulate and accomplish various downstream tasks based on prompt learning.

\textbf{Classification Task}. Classification tasks usually require classifying sentences or sentence pairs into different classes. This means that each class expresses a different meaning. We define a text classification dataset as $D=\{(x_n,y_n)\}_{n-1}^{N}$, where $x$ is the $nth$ sample (a total of $N$ samples). The $y_n$ is the label of $x_n$, and $y_n\in \mathcal{Y}^{N\times M}$, $M$ is the number of classes. Each $y_n$ can be mapped to $y^{(m)}$, and  $y^{(m)}$ can be mapped to the label word  $\mathcal{LW}^{(m)}$, and then $\mathcal{LW}^{(m)}$  and other prompt words together form the template $t^{(m)}\in\mathcal{T}^{m}$. The input to the model can be expressed as:
\begin{equation}\label{key}
x_{input}={\rm [CLS]}{\bm t}^{(m)}{\rm[SEP]}x_n{\rm[EOS]}
\end{equation}

The probability that the model predicts the sample $x_n$ to be the label $y^{(m)}$ is:
\begin{equation}\label{key}
\small
P(y^{(m)}|x_n)=\dfrac{(1-P(\mathcal{LW}^{(m)}={\rm Rep}|[t^{(m)},x_n]))}{\sum_{m=1}^{M}(1-P(\mathcal{LW}^{(m)}={\rm Rep}|[t^{(m)},x_n]))}
\end{equation}
where $M$ is the number of categories, {\rm Rep} is the abbreviation for {\rm Replaced}, and then the predicted label $\hat{y}$ for sample $x_n$ is:
\begin{equation}\label{key}
\hat{y}={\rm max}(P(y^{(1)}|x_n),P(y^{(2)}|x_n),...,P(y^{(M)}|x_n))
\end{equation}

\textbf{Regression Task}. The most significant difference between regression tasks and classification tasks is that the label $y$ is in a bounded interval $[V_1, V_2]$. Because labels are continuous values, we cannot map to various label words. In fact, we only need a label word $\mathcal{LW}$, and the model will predict the probability that the label word is replaced. $\mathcal{LW}$ and other prompt words together form the template $t$. We just need to map the probability distribution $[0, 1]$ to the label space $[V_1, V_2]$. The input to the model can be expressed as:
\begin{equation}\label{key}
x_{input}={\rm [CLS]}{\bm t}{\rm[SEP]}x_n{\rm[EOS]}
\end{equation}
The model finally predicts that the value of $\hat{y}$ is:
\begin{equation}\label{key}
\hat{y}=|V_2-V_1|\ast P(\mathcal{LW}={\rm Rep}|[t,x_n])) + V_1
\end{equation}

\section{Experiments}
\subsection{Datasets}
In this paper, we select 15 popular NLP datasets to evaluate our proposed method.  For SNLI~\cite{bowman2015large_snli} and datasets from GLUE~\cite{wang2019glue}, including SST-2~\cite{socher2013recursive_sst-2}, CoLA~\cite{warstadt2019neural_cola}, MNLI~\cite{williams2018broad_mnli}, QNLI~\cite{rajpurkar2016squad}, RTE~\cite{dagan2005pascal_rte1,bar2006second,giampiccolo2007third_rte3,bentivogli2009fifth_rte4}, MRPC~\cite{dolan2005automatically_mrpc}, QQP\footnote{\url{https://www.quora.com/q/quoradata/}} and STS-B~\cite{cer2017semeval_sts-b}, we follow the work~\citep{zhang2020revisiting} and use their original dev sets for testing. For MR~\cite{pang2005seeing_mr}, CR~\cite{hu2004mining_cr}, MPQA~\cite{wiebe2005annotating_mpqa}, Subj~\cite{pang2004sentimental_subj}, we follow the research~\citep{gao2021making} and simply randomly sample 2,000 examples as the testing set and leave them out from training. For SST-5~\cite{socher2013recursive_sst-2} and TREC~\cite{voorhees2000building_trec}, we use official test sets. 
\subsection{Experimental Setup}
To evaluate the proposed method, we compare the performance of (1) MLM-BERT (mlm-based prompt learning), (2) NSP-BERT (nsp-based prompt learning), (3) MLM-RoBERTa (mlm-based prompt learning) and our (4) RTD-ELECTRA (rtd-based prompt learning) models for zero-shot learning. We compare base-sized (110M) and large-sized (335M) models. All models do not use any training set and development set, only the test set. The experimental code in this paper is based on bert4keras\footnote{\url{https://github.com/bojone/bert4keras}}. In this paper, we complete the prompt learning based on manual templates. The templates and label words used in the experiment are listed in Table \ref{t1}.
Note that for the STS-B regression task, our predicted label word is: \texttt{NO}. The hardware of experiments is a GPU: RTX 3090.

\subsection{Main Experimental Results}
We compare the zero-shot learning performance of RTD-ELECTRA with MLM-BERT, NSP-BERT and MLM-RoBERTa. The experimental results are shown in Table \ref{t2}. First, all prompt-based zero-shot prediction achieves much better performance than the \textit{majority class}, showing the pre-encoded knowledge in pre-training models. For the base-sized models, RTD-ELECTRA$_{base}$ achieves an average score of 52.39\%, which is \textbf{3.51\%} higher than MLM-RoBERTa$_{base}$, \textbf{7.84\%} higher than NSP-BERT$_{base}$, and \textbf{9.25\%} higher than MLM-BERT$_{base}$. For the large-sized models, RTD-ELECTRA performs better than base-sized RTD-ELECTRA. On all 15 NLP datasets, RTD-ELECTRA$_{large}$ performs the best on 12 datasets. Except on SST-5 dataset, RTD-ELECTRA$_{large}$ is 0.1\% lower than MLM-RoBERTa, and RTD-ELECTRA$_{large}$ is lower than NSP-BERT$_{large}$ on STS-B and MPQA datasets. For average performance on all datasets, our RTD-ELECTRA$_{large}$ achieves about \textbf{8.22\%}, \textbf{12.97\%} and \textbf{13.63\%} improvement in zero-shot performance compared to MLM-RoBERTa$_{large}$, NSP-BERT$_{large}$ and MLM-BERT$_{large}$, respectively. In particular, RTD-ELECTRA$_{large}$ achieves an astonishing \textbf{90.1\%} accuracy on the SST-2 dataset without any training data. From the above experimental results, the zero-shot learning performance of the ELECTRA model is excellent, mainly due to its unique RTD pre-training task. 

\subsection{Analysis of Various Prompts}
To analyze the impact of different prompts on the results, we compared the zero-shot performance of four prompt-based baselines on the SST-2 task with various prompts. The experimental results are shown in Table \ref{t3}, and we can see that the zero-shot performance of our proposed RTD-ELECTRA model is significantly higher than other baseline models under all four different prompts. We can find that the MLM-RoBERTa and RTD-ELECTRA models are generally better than the MLM-BERT and NSP-BERT models, which also shows that the pre-training methods of RoBERTa and ELECTRA are indeed better than the original BERT model. In addition, although the zero-shot performance of the ELECTRA model fluctuates for different prompts, the performance is also satisfactory compared to BERT and RoBERTa. 
\begin{table}[t]
	\renewcommand\arraystretch{1.3}
	\setlength\tabcolsep{1.25pt}
	\small
	\begin{tabular}{l|l|c}
		\toprule[1pt]
		Prompts & Models & Acc. \\ \hline
		\multirow{4}{*}{\begin{tabular}[c]{@{}l@{}}$\mathcal{P}_1$:\\{[}Input{]} \textless{}SEP\textgreater~It was {[}label{]}.\\ Label words: \textit{great/terrible}\end{tabular}} & MLM-BERT & 60.1 \\ \cline{2-3} 
		& NSP-BERT & 60.7 \\ \cline{2-3} 
		& MLM-RoBERTa & 83.6 \\ \cline{2-3} 
		& RTD-ELECTRA & \textbf{87.4} \\ \hline
		\multirow{4}{*}{\begin{tabular}[c]{@{}l@{}}$\mathcal{P}_2$:\\{[}Input{]} \textless{}SEP\textgreater ~It was {[}label{]}.\\ Label words: \textit{good/bad}\end{tabular}} & MLM-BERT & 54.5 \\ \cline{2-3} 
		& NSP-BERT & 62.2 \\ \cline{2-3} 
		& MLM-RoBERTa & 77.6 \\ \cline{2-3} 
		& RTD-ELECTRA & \textbf{84.6} \\ \hline
		\multirow{4}{*}{\begin{tabular}[c]{@{}l@{}}$\mathcal{P}_3$:\\{[}Input{]} \textless{}SEP\textgreater ~This movie is {[}label{]}.\\ Label words: \textit{great/terrible}\end{tabular}} & MLM-BERT & 68.3 \\ \cline{2-3} 
		& NSP-BERT & 67.9 \\ \cline{2-3} 
		& MLM-RoBERTa & 84.6 \\ \cline{2-3} 
		& RTD-ELECTRA & \textbf{86.5} \\ \hline
		\multirow{4}{*}{\begin{tabular}[c]{@{}l@{}}$\mathcal{P}_4$:\\{[}Input{]} \textless{}SEP\textgreater ~This movie is {[}label{]}!\\ Label words: \textit{great/terrible}\end{tabular}} & MLM-BERT & 66.5 \\ \cline{2-3} 
		& NSP-BERT & 65.1 \\ \cline{2-3} 
		& MLM-RoBERTa & 83.5 \\ \cline{2-3} 
		& RTD-ELECTRA & \textbf{86.6} \\ 
		\bottomrule[1pt]
	\end{tabular}
	\caption{Zero-shot results with different prompts on SST-2 task (Large-sized models).}
	\label{t3}
\end{table}
\begin{table}[t]
	\centering
	\small 
	\renewcommand\arraystretch{1.3}
	\setlength\tabcolsep{1pt}
	\begin{tabular}{lcccccc}
		\toprule[1pt]
		\textbf{Methods} & \textbf{\begin{tabular}[c]{@{}c@{}}SST-2\\ (acc.)\end{tabular}} & \textbf{\begin{tabular}[c]{@{}c@{}}MR\\ (acc.)\end{tabular}} & \textbf{\begin{tabular}[c]{@{}c@{}}MPQA\\ (acc.)\end{tabular}} & \textbf{\begin{tabular}[c]{@{}c@{}}MNLI\\ (m-acc.)\end{tabular}} \\ \hline 
		\multicolumn{5}{c}{RTD-Prompt and Predict (Zero-Shot)} \\ \hline
		ELECTRA-small & 66.7 & 62.4 & 60.8 & 40.0 \\
		ELECTRA-base & 81.0 & 85.2 & 61.7 & 52.6 \\
		ELECTRA-large & \textbf{90.1} & \textbf{84.6} & \textbf{68.2} & \textbf{56.1}  \\ \hline
		\multicolumn{5}{c}{Pre-train, Fine-Tune (Few-Shot)} \\ \hline
		ELECTRA-small & 57.5 (2.4) & 52.2 (1.1) & 57.3 (3.3) & 34.6 (1.8)  \\
		ELECTRA-base & 75.7 (3.3) & 57.3 (2.1) & 56.9 (1.4) & \textbf{38.5} (2.0) \\
		ELECTRA-large & \textbf{77.9} (2.6) & \textbf{72.0} (0.8) & \textbf{58.8} (5.0) & 37.6 (3.0)  \\
		\bottomrule[1pt]
	\end{tabular}
	\caption{Performance of RTD-prompt prediction and fine-tuning on different sized models. We use $K=16$ (per class) for fine-tuning and report mean (and standard deviation) performance over five random training set splits for few-shot experiments.}
	\label{t4}
\end{table}
\begin{figure*}[t]
	\centering
	\includegraphics[width=1\linewidth]{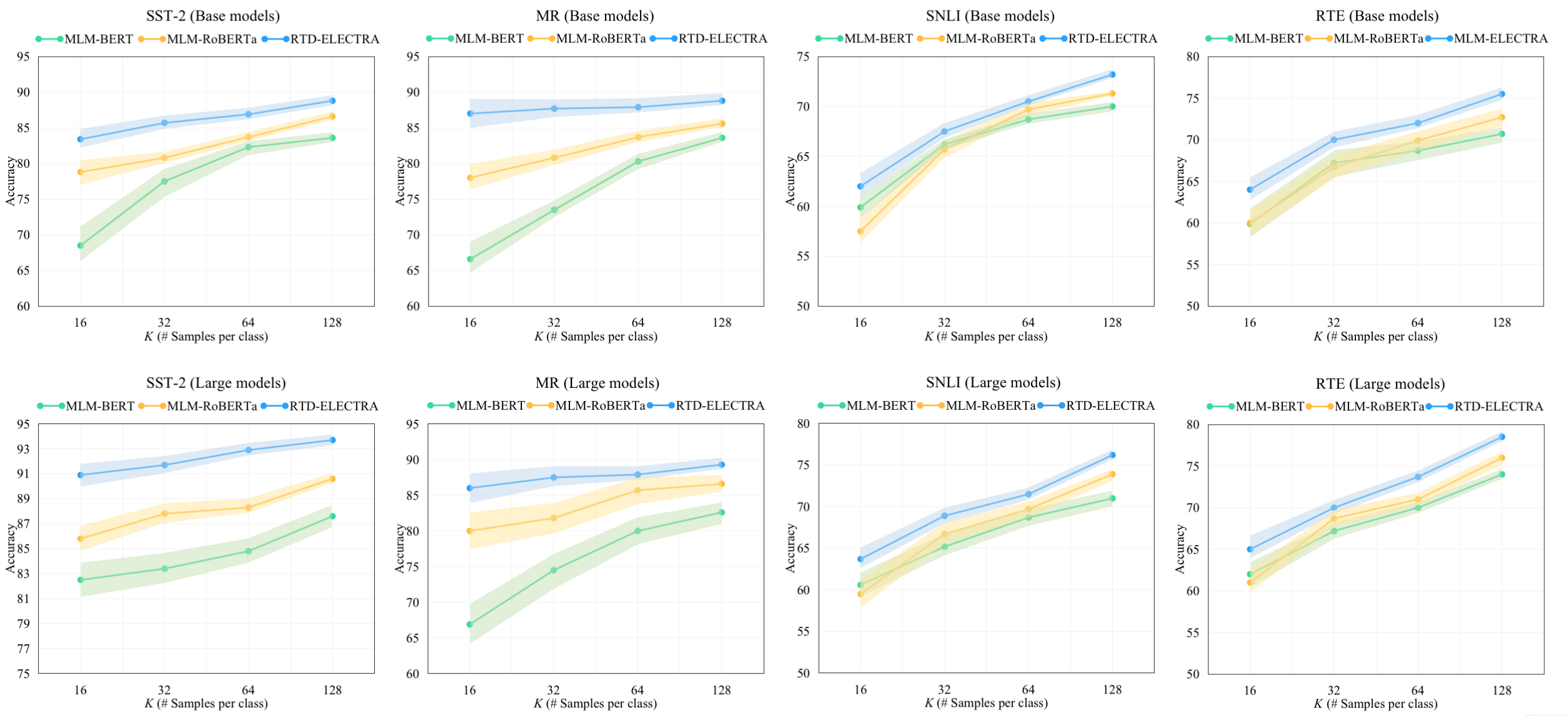}
	\caption{ Few-shot performance of MLM-BERT, MLM-RoBERTa and
our RTD-ELECTRA with prompt-based fine-tuning
as $K$ (number of samples per label) samples. 
	}
	\label{f4}
\end{figure*}
\subsection{RTD-Prompt Prediction VS. Fine-Tuning}
In this work, we test the small (14M), base (110M) and large (335M) versions of the ELECTRA\footnote{\url{https://github.com/google-research/electra}} model. We report the zero-shot performance of the models and the fine-tuning performance of the models under few-shot. All experimental results are shown in Table \ref{t4}. We find that the zero-shot performance of all prompt-based RTD-ELECTRA is better than the few-shot performance of fine-tuning-based ELECTRA. This shows the advantage of RTD-prompt learning under zero-shot and few-shot. In addition, it can also be found from the experimental results that the size of the model has a significant influence on the prompt-based method but has little effect on the fine-tuning-based method. This is because the model's size largely determines the amount of knowledge learned in the pre-training process. Generally, the larger the model, the better the effect of prompt learning. For fine-tuning, larger models are not necessarily better, and larger models may be more challenging to converge with insufficient training data.

\subsection{Study of Prompt Few-Shot Learning}
We compare our approach (RTD-ELECTRA) with two few-shot learning approaches based on pre-trained masked language models (MLM-BERT and MLM-RoBERTa). Compared with zero-shot learning, few-shot learning uses a small number of samples to continue training the model based on the pre-training task (e.g., MLM, RTD), which means that the model's parameters are updated. Figure \ref{f4} shows RTD-prompt (ELECTRA) fine-tuning and MLM-prompt fine-tuning (BERT and RoBERTa) performance as the number of instances ($K$) increases on four datasets. We evaluate both base-sized and large-sized models. From the experimental results, we can see that when $K=\{16, 32, 64, 128\}$, the performance of ELECTRA is always higher than that of BERT and RoBERTa. Moreover, for the binary sentiment classification task, RTD-prompt fine-tuning is significantly better than MLM-prompt fine-tuning, and it is more stable. When performing the RTD pre-training task, the discriminator of ELECTRA handles a binary classification task ( original/replaced). Therefore, the RTD task is more similar to the binary sentiment classification task than the MLM task.

\section{Conclusion}
In this paper, we explore the ELECTRA model for prompt-based zero-shot learning on NLP tasks. We propose a novel RTD-based prompt learning method. Through extensive experiments on 15 various datasets, we find that the ELECTRA model performs surprisingly well as a zero-shot and few-shot learner, proving the ELECTRA model has more potential to be stimulated. For instance, our RTD-ELECTRA$_{large}$ achieves an astonishing 90.1\% zero-shot performance on SST-2 task. The superior performance is mainly due to the well-designed RTD pre-training task, enabling it to learn more pre-knowledge.  Our work proves that ELECTRA is an excellent zero-shot learner, Too. Prompt-based learning can better connect pre-training tasks and various downstream tasks, and we believe that the design of model pre-training tasks will be an essential research direction. In addition, we will also try to use automatic or continuous prompts for ELECTRA in the future.

\bibliography{anthology,custom}
\bibliographystyle{acl_natbib}

\clearpage
\end{document}